\pdfoutput=1
\documentclass[11pt]{article}

\usepackage[]{acl}

\usepackage{times}
\usepackage{latexsym}
\usepackage{multicol}
\usepackage{multirow}
\usepackage{graphicx}

\usepackage{amsmath}

\usepackage[T1]{fontenc}
\usepackage[utf8]{inputenc}

\usepackage{microtype}

\usepackage{inconsolata}

\title{Co-Trained Retriever-Generator Framework for Question Generation in Earnings Calls}

\author{Yining Juan,\textsuperscript{\rm 1}
    Chung-Chi Chen,\textsuperscript{\rm 2}
    Hen-Hsen Huang,\textsuperscript{\rm 3}
    Hsin-Hsi Chen\textsuperscript{\rm 1} \\
   \textsuperscript{\rm 1} Department of Computer Science and Information Engineering, \\ 
   National Taiwan University, Taiwan \\
    \textsuperscript{\rm 2} AIST, Japan \\
    \textsuperscript{\rm 3}
    Institute of Information Science, Academia Sinica, Taiwan \\
    ynjuan@nlg.csie.ntu.edu.tw,
    c.c.chen@acm.org,\\
    hhhuang@iis.sinica.edu.tw,
    hhchen@ntu.edu.tw}

\begin{document}
\maketitle
\begin{abstract}
In diverse professional environments, ranging from academic conferences to corporate earnings calls, the ability to anticipate audience questions stands paramount. Traditional methods, which rely on manual assessment of an audience's background, interests, and subject knowledge, often fall short—particularly when facing large or heterogeneous groups, leading to imprecision and inefficiency. While NLP has made strides in text-based question generation, its primary focus remains on academic settings, leaving the intricate challenges of professional domains, especially earnings call conferences, underserved. Addressing this gap, our paper pioneers the multi-question generation (MQG) task specifically designed for earnings call contexts. Our methodology involves an exhaustive collection of earnings call transcripts and a novel annotation technique to classify potential questions. Furthermore, we introduce a retriever-enhanced strategy to extract relevant information. With a core aim of generating a spectrum of potential questions that analysts might pose, we derive these directly from earnings call content. Empirical evaluations underscore our approach's edge, revealing notable excellence in the accuracy, consistency, and perplexity of the questions generated.
\end{abstract}

\section{Introduction}
\label{sec:introduction}

Oral presentations are ubiquitous in professional settings. Researchers at conferences must convey their findings effectively, while politicians prepare meticulously for debates during election cycles. Corporate leaders, too, need to clearly communicate operational updates to stakeholders. A critical aspect of these presentations is anticipating potential audience questions.
Traditionally, presenters predict questions by assessing the audience's background, interests, and subject familiarity. However, this method can be laborious and imprecise, particularly with diverse or large audiences. Moreover, audiences may pose unexpected questions. The field of Natural Language Processing (NLP) has extensively explored question generation from texts, as evidenced by the Stanford Question Answering Dataset (SQuAD) \cite{rajpurkar-etal-2016-squad} and Google's Natural Questions (NQ) \cite{kwiatkowski2019natural}. These datasets primarily focus on factual information retrieval from Wikipedia-like sources \cite{rajpurkar-etal-2016-squad, kwiatkowski2019natural}. Most question generation studies target educational settings, creating questions from textbooks for student self-assessment \citep{heilman2009question, le2014automatic, mostow2009generating}. While beneficial in education, these methods may not translate well to professional contexts where questions are complex and require domain expertise.

Research on automatic question generation for professional domains is limited. An example is the study by Chen et al. \cite{chen-etal-2021-finqa}, which generated question-answer pairs from financial documents, focusing more on factual extraction than preparing presenters for audience questions. Given this background, there is a clear need for auto-generated questions suited for professional settings, where nuanced and specialized audience inquiries pose unique challenges. Earnings calls in corporate settings exemplify such scenarios. In these calls, executives discuss financial updates with investors and must be prepared to answer a wide array of questions, ranging from fiscal stability to future strategies, with significant impact on corporate image and investor decisions.

This paper follows a previous study \cite{juan-etal-2023-generating} and focuses on earnings conference calls (ECC) for exploring the task of multi-question generation (MQG). MQG aims to generate a diverse set of potential questions from earnings call content, differing from traditional one-to-one question generation tasks \cite{du-etal-2017-learning,song-etal-2018-leveraging}. This one-to-many question generation approach more accurately reflects real-world scenarios, helping to better understand audience perspectives and prepare for their inquiries.
To address the research challenges, we compile a comprehensive dataset of earnings call transcripts. We also develop a unique annotation strategy to categorize analyst-posed questions, enhancing the depth and relevance of generated questions. We propose a retriever-enhanced question generation framework using the LLaMA framework \cite{touvron2023llama} combined with LoRA adapters \cite{hu2021lora}, employing dual models as retrievers and generators. The retriever selects relevant paragraphs from a presentation, which are then used by the generator to create reference questions. The quality of these questions informs adjustments to the retriever's parameters, improving paragraph selection.

Our empirical evaluation employs various automatic metrics and scenarios, including random and BM25 retriever configurations. We also conduct an ablation analysis and evaluate the use of question types as control codes for the generator and the impact of retriever parameter magnitude.
The contributions of this paper are:
(1) Presenting an innovative retriever-augmented question generation framework, which uses question types as guides to produce high-quality questions.
(2) Demonstrating that our approach outperforms existing methods in generating accurate, varied, and clear questions.

\section{Related Work}

MQG in the context of ECCs presents unique challenges compared to conventional approaches. Traditional models typically rely on both context and corresponding answers to generate questions \cite{du-etal-2017-learning,song-etal-2018-leveraging}. However, in ECCs, when assisting presenters in preparing their reports, only the presentation content is available, necessitating the model to independently generate a range of questions without answer-based guidance. This requires the model to identify potential areas of interest or relevance within the presentation.
Another challenge is the complex nature of questions posed by financial analysts during earnings calls. In contrast to the more straightforward questions in datasets like SQuAD or Natural Questions, questions from financial analysts often include contextual introductions and may encompass nested or multiple inquiries spanning various topics \cite{rajpurkar-etal-2016-squad, kwiatkowski2019natural}. Traditional question generation focuses on producing single, isolated questions.

To effectively automate the generation of interconnected questions, a model must not only comprehend the extensive and complex content of ECCs but also emulate the intricate reasoning typical of a financial analyst. This task extends beyond basic content analysis, striving to replicate the cognitive depth and expertise characteristic of experienced analysts.

% --------------- Table ---------------
\begin{table}[t]
    \centering
    \resizebox{\columnwidth}{!}{
    \begin{tabular}{c|c|c|c|c}
    
    \# Transcripts & \# Questions & \begin{tabular}[c]{@{}c@{}}Avg. \\ Presentation \\ Length\end{tabular} & \begin{tabular}[c]{@{}c@{}}Avg. \\ Question \\ Length\end{tabular} & \begin{tabular}[c]{@{}c@{}}Avg.\\ \# Questions\\ per transcripts\end{tabular} \\ \hline
    5,999          & 52,889       & 3,582                                                                   & 82                                                                  & 9                                                                             \\ 
    \end{tabular}
    }
        \caption{Statistics of the ECCT dataset.}
    \label{tab:dataset_statistic}
\end{table}
% --------------- Table ---------------

\section{Dataset Creation}
We harvested publicly accessible earnings call transcriptions (ECCTs) for 300 companies listed in the S\&P 500 index from SeekingAlpha.\footnote{https://seekingalpha.com/} Spanning from January 2004 to February 2023, our collection aggregates 5,999 ECCTs. Intrinsically, an ECCT can be partitioned into two principal segments: the ``Prepared Remarks'' - an exposition of the enterprise's fiscal metrics for the discussed interval, and a subsequent ``Question and Answer'' section. 
Our data processing results are summarized in Table~\ref{tab:dataset_statistic}. Notably, 95\% of our question dataset fell within 178 words, with the longest presentation recorded at 18,923 words.

\section{Co-Trained
Retriever-Generator Framework}
Given the extensive content of oral presentations, most transformer-based models encounter token constraints. Recognizing that audience questions typically target presentation specifics, we introduce a Co-Trained Retriever-Generator Framework. 
During training, the Retriever identifies relevant presentation segments, guiding the Generator to craft audience-oriented questions. Improved retrieval boosts generation accuracy, with the consequent loss iteratively optimizing both models, enhancing the Retriever's unsupervised capability.

Formally, we define a segmentation function, \(f(P)\), for a typical presentation \(P\). This function divides \(P\) into \(n\) segments, \(p_{1:n}\), with \(P = \text{concat}(p_{1:n})\) representing the continuous combination of these segments. Each segment, \(p_i\), adheres to a 128-token limit. We then select a subset of these segments, combining them into a single cohesive input segment, \(p_{\text{in}} = \text{concat}(p_{i:j})\). Our generative model uses \(p_{\text{in}}\) to produce a query \(q\), formulated as \(q = g(p_{\text{in}})\).

Considering the complexity of retrieving manager presentation passages that are highly relevant to an analyst's query, a deep understanding of both the specialized financial context and the intent of the analyst's question is crucial. To effectively tackle this challenge, we implement a retriever-generator model, based on the LLaMa architecture~\cite{touvron2023llama}. 
The retriever \( R \), named Prompt-Based Retriever (ProRetriever), and generator \( G \) models calculate a relevance metric through a text-to-text framework for a given question \( q \) and presentation snippet \( p \). 

Our crafted prompt reads:\

\noindent  \textit{Given a manager's presentation transcript during an earnings call and an analyst's query, discern if the query is deeply anchored, tangentially connected, or aloof from the manager's discourse? (``Highly Related''/``Partially Related''/``Not Related'') Transcript: ${presentation}$ Question: ${question}$ Assistant: The assessment is [MASK]"}

\( R \) assesses the alignment of the [MASK] token with labels ``Highly Related'', ``Partially Related'', or ``Not Related'', computing the relevance score for each paragraph as: 
\[
\small
\text{Score}(p, q) = \text{P}(\text{``Highly''}) + \text{P}(\text{``Partially''}) - \text{P}(\text{``Not''}).
\]
\( R \) then selects the top-\( k \) paragraphs, \( p^{(1)}, p^{(2)}, \ldots, p^{(k)} \), in their presentation sequence. These paragraphs are forwarded to \( G \) for further processing. If \( r \) is the generated output and \( r^* \) the standard answer, the generator's cross-entropy loss is calculated as:
\[
L_{\text{gen}} = -\sum_{t} r^*_t \log p(r_t | p^{(1)}, p^{(2)}, \ldots, p^{(k)}),
\]
where \( p(r_t | p^{(1)}, p^{(2)}, \ldots, p^{(k)}) \) represents the likelihood of generating token \( r_t \) given the retrieved paragraphs and previous tokens. Through this loss, we adjust the parameters for both retriever and generator, iteratively optimizing the system's performance.

Notably, the Retriever is only utilized during the training phase. After training, users simply select the segments of interest from a presentation and input them into the trained generator. This process yields questions that the generator predicts an analyst is likely to ask.

\section{Experiment}
\subsection{Experimental Setup}
Our system utilized the Alpaca-Lora-7B~\cite{alpaca, hu2021lora} as the question generator and the Guanaco-33B~\cite{dettmers2023qlora} for retrieving relevant passages. Both models, derivatives of the open-source LLaMA, were optimized using quantization and LoRA techniques. This allowed fine-tuning of only a small fraction of parameters: 0.24\% for the generator and 1.24\% for the retriever. We applied 8-bit and 4-bit quantization to the generator and retriever, respectively, and used gradient checkpointing for efficient VRAM utilization. The input length limits were set to 1400 tokens for the generator and 512 for the retriever, with the generator's output capped at 200 tokens.\footnote{Please refer to Appendix~\ref{sec:Implement Details} for hyperparameter settings.}

\begin{table*}
\centering
\small
\begin{tabular}{ll|c|c|c|c|c|c}
    \multirow{2}{*}{Generator} & \multirow{2}{*}{Retriever}         & \multicolumn{5}{c|}{Correctness}                                                           & Diversity  \\
           &   & BLEU-4          & ROUGE-2         & ROUGE-L          & METEOR           & BERTScore        & Sem-Ent                     \\ 
\hline
GPT-4  & -       & 1.347           & 4.821           & 22.576           & 16.329           & 79.115           & 1.711                       \\ 
\hline
\multirow{4}{*}{Alpaca-Lora}  & -   & 0.918           & 4.987           & 21.633           & 12.704           & 76.866           & 1.706                       \\ 
& Random       & 2.039           & 6.474           & 27.287           & 20.428           & 77.976           & 1.717                       \\ 
& BM25         & 2.025           & 6.971           & 27.931           & 20.082           & 79.024           & 1.728                       \\ 
& ProRetriever & \textbf{2.389}* & \textbf{7.255}* & \textbf{28.891}* & \textbf{22.063}* & \textbf{81.566}* & \textbf{1.759*}            
\end{tabular}
\caption{Evaluation from correctness and diversity aspects. * denote that the improvement over the best-performing baseline (BM25) is statistically significant under t-test with p-value $<$ 0.05.}
\label{tab: Evaluation results from correctness and diversity aspects}
\end{table*}

To assess the efficacy of our Co-Trained Retriever-Generator Framework, we compared it against four different retriever techniques: the Random Retriever and the BM25 Retriever. This comparison aimed to demonstrate how our retrieve-augmented approach enhances the question generation capabilities of the model. Additionally, for the generator component, we conducted comparisons with a zero-shot pretrained Alpaca-Lora model and GPT-4 to further validate the effectiveness of our method. 
\textbf{Random Retriever:} For each reference question, this method arbitrarily selected ``k'' presentation passages, creating an input paragraph for the generator.
\textbf{BM25 Retriever:} The BM25 algorithm replaced random selection, picking the top-k pertinent passages relative to each reference question. The resultant paragraphs, when paired with their associated reference questions, trained the generator.

Regardless of the retrieval method used in training, our study evaluated these strategies by varying the retrieval approach during testing. This allowed us to observe how different training retrieval techniques affected performance when applied with different retrieval methods in testing. This approach helped clarify how the choice of retrieval method impacts the overall effectiveness of the model.

\subsection{Evaluation Metric}
\label{sec: Evaluation Metric}
Several metrics facilitate our exploration of correctness:
BLEU~\cite{papineni2002bleu}, ROUGE~\cite{lin2004rouge}, METEOR~\cite{banerjee2005meteor}, and BERTScore~\cite{zhang2019bertscore}.
Given that the LLM typically produces text with human-like readability and rarely exhibits phrase repetition degeneration, our evaluation of Diversity shifted. Instead of assessing n-gram overlap, we now focus on the model's ability to simulate diverse questions from different analysts' perspectives, themes, and sections of the same presentation. This change reflects our aim to evaluate the model's capacity for generating a wide range of queries based on a single presentation.
Inspired by the method proposed in the previous study\cite{han2022measuring}, we evaluate the semantic diversity of generated questions (Sem-Ent) from an entropy perspective.\footnote{Please refer to Appendix~\ref{sec:Details of Sem-Ent} for details.}

\subsection{Automatic Evaluation}
We utilized our trained retriever, as detailed in Table~\ref{tab: Evaluation results from correctness and diversity aspects}. This retriever identifies relevant presentation segments based on the reference questions. Then, models, trained under various retrieval methodologies, utilize these segments to generate questions, allowing us to derive correctness and diversity metrics.

It is evident from the results that our method demonstrates superior performance in terms of correctness when compared to models developed using alternative retrieval techniques. Upon comparing with the baseline models, it becomes evident that under the retrieve-augment setup, both Correctness and Diversity performance metrics have seen a substantial improvement over the methodologies employed in the pilot study.

\subsection{Human Evaluation}
We employed three annotators with financial expertise to evaluate the generated questions in two dimensions: (1) Logic and Consistency (LC) and (2) Professionalism (PF). This assessment also included a comparison with questions asked by professional analysts and a BM-25 baseline, to highlight differences between generated and professional questions.
For LC, a score of 4 represents a perfect question in both dimensions; 3 indicates a minor issue in one dimension; 2 signifies minor issues in both dimensions; and 1 denotes major issues in any dimension. For PF, a score of 3 corresponds to a critical question; 2 to a reasonable question; and 1 indicates a lack of professionalism.

Table~\ref{tab:Human Evaluation} presents the evaluation results, representing the average scores from three annotators for 50 randomly selected questions. In terms of logic and consistency, the results show that the models perform well, achieving scores comparable to those of professionals. However, in terms of professionalism, questions generated using the proposed ProRetriever model are more reasonable than those generated using BM25, but they do not attain the critical level achieved by professionals. These findings suggest that future research in the MQG task should focus on subjective but crucial aspects such as professionalism.

\begin{table}[t]
    \centering
    \resizebox{\columnwidth}{!}{
    \begin{tabular}{l|cc}
    & \textbf{Logic and Consistency } & \textbf{Professionalism} \\
    \hline
    BM25  & 3.73 & 1.86 \\
    ProRetriever & 3.75 & 2.05 \\
    \hline
    Analyst & 3.79 & 2.25  \\
    \end{tabular}
    }
        \caption{Human Evaluation.}
    \label{tab:Human Evaluation}
\end{table}

\section{Conclusion}
This study explores the nascent area of Multi-Question Generation (MQG), specifically within the context of earnings call conferences. We have developed a specialized retriever-enhanced strategy for relevant paragraph retrieval, which has proven effective in generating diverse and contextually appropriate questions.
Human evaluation results highlight the current gap between the models' question generation capabilities and those of professionals. We plan to release the dataset and code to serve as a foundation for future research in this field.
In subsequent phases, our objective is to broaden the application of our methods to a variety of presentation settings, including academic conferences, debates, and thesis defenses.

\section*{Limitation}
First, our method is domain-specific, focusing solely on earnings call transcripts. This narrow scope may limit the model's generalizability to other professional settings, such as academic presentations, political debates, or technical product launches, where the nature and complexity of questions can differ significantly. Extending this approach to broader contexts would require domain adaptation or transfer learning, which could introduce additional challenges.
Second, while our retriever-enhanced framework improves question diversity and correctness, it still relies heavily on the availability of extensive training data. The quality of generated questions could degrade in settings where transcripts or previous interactions are sparse, leading to a lack of depth or relevance in the output.
Lastly, our approach is computationally intensive, particularly during training due to the use of large-scale transformer models and dual model frameworks for retriever and generator. This may pose practical limitations in terms of accessibility and deployment for organizations with limited computational resources.

Addressing these limitations in future work could improve the model's versatility, efficiency, and real-world applicability across a wider array of professional domains.

\bibliography{anthology,custom}

\begin{thebibliography}{18}
\expandafter\ifx\csname natexlab\endcsname\relax\def\natexlab#1{#1}\fi

\bibitem[{Banerjee and Lavie(2005)}]{banerjee2005meteor}
Satanjeev Banerjee and Alon Lavie. 2005.
\newblock Meteor: An automatic metric for mt evaluation with improved correlation with human judgments.
\newblock In \emph{Proceedings of the acl workshop on intrinsic and extrinsic evaluation measures for machine translation and/or summarization}, pages 65--72.

\bibitem[{Chen et~al.(2021)Chen, Chen, Smiley, Shah, Borova, Langdon, Moussa, Beane, Huang, Routledge, and Wang}]{chen-etal-2021-finqa}
Zhiyu Chen, Wenhu Chen, Charese Smiley, Sameena Shah, Iana Borova, Dylan Langdon, Reema Moussa, Matt Beane, Ting-Hao Huang, Bryan Routledge, and William~Yang Wang. 2021.
\newblock \href {https://doi.org/10.18653/v1/2021.emnlp-main.300} {{F}in{QA}: A dataset of numerical reasoning over financial data}.
\newblock In \emph{Proceedings of the 2021 Conference on Empirical Methods in Natural Language Processing}, pages 3697--3711, Online and Punta Cana, Dominican Republic. Association for Computational Linguistics.

\bibitem[{Dettmers et~al.(2023)Dettmers, Pagnoni, Holtzman, and Zettlemoyer}]{dettmers2023qlora}
Tim Dettmers, Artidoro Pagnoni, Ari Holtzman, and Luke Zettlemoyer. 2023.
\newblock Qlora: Efficient finetuning of quantized llms.
\newblock \emph{arXiv preprint arXiv:2305.14314}.

\bibitem[{Du et~al.(2017)Du, Shao, and Cardie}]{du-etal-2017-learning}
Xinya Du, Junru Shao, and Claire Cardie. 2017.
\newblock \href {https://doi.org/10.18653/v1/P17-1123} {Learning to ask: Neural question generation for reading comprehension}.
\newblock In \emph{Proceedings of the 55th Annual Meeting of the Association for Computational Linguistics (Volume 1: Long Papers)}, pages 1342--1352, Vancouver, Canada. Association for Computational Linguistics.

\bibitem[{Han et~al.(2022)Han, Kim, and Chang}]{han2022measuring}
Seungju Han, Beomsu Kim, and Buru Chang. 2022.
\newblock Measuring and improving semantic diversity of dialogue generation.
\newblock \emph{arXiv preprint arXiv:2210.05725}.

\bibitem[{Heilman and Smith(2009)}]{heilman2009question}
Michael Heilman and Noah~A Smith. 2009.
\newblock Question generation via overgenerating transformations and ranking.
\newblock \emph{DTIC Document}.

\bibitem[{Hu et~al.(2021)Hu, Shen, Wallis, Allen-Zhu, Li, Wang, Wang, and Chen}]{hu2021lora}
Edward~J Hu, Yelong Shen, Phillip Wallis, Zeyuan Allen-Zhu, Yuanzhi Li, Shean Wang, Lu~Wang, and Weizhu Chen. 2021.
\newblock Lora: Low-rank adaptation of large language models.
\newblock \emph{arXiv preprint arXiv:2106.09685}.

\bibitem[{Juan et~al.(2023)Juan, Chen, Huang, and Chen}]{juan-etal-2023-generating}
Yining Juan, Chung-Chi Chen, Hen-Hsen Huang, and Hsin-Hsi Chen. 2023.
\newblock \href {https://doi.org/10.18653/v1/2023.inlg-main.35} {Generating multiple questions from presentation transcripts: A pilot study on earnings conference calls}.
\newblock In \emph{Proceedings of the 16th International Natural Language Generation Conference}, pages 449--454, Prague, Czechia. Association for Computational Linguistics.

\bibitem[{Kwiatkowski et~al.(2019)Kwiatkowski, Palomaki, Redfield, Collins, Parikh, Alberti, Epstein, Polosukhin, Devlin, Lee et~al.}]{kwiatkowski2019natural}
Tom Kwiatkowski, Jennimaria Palomaki, Olivia Redfield, Michael Collins, Ankur Parikh, Chris Alberti, Danielle Epstein, Illia Polosukhin, Jacob Devlin, Kenton Lee, et~al. 2019.
\newblock Natural questions: a benchmark for question answering research.
\newblock \emph{Transactions of the Association for Computational Linguistics}, 7:453--466.

\bibitem[{Le et~al.(2014)Le, Kojiri, and Pinkwart}]{le2014automatic}
Nguyen-Thinh Le, Tomoko Kojiri, and Niels Pinkwart. 2014.
\newblock Automatic question generation for educational applications--the state of art.
\newblock In \emph{Advanced Computational Methods for Knowledge Engineering: Proceedings of the 2nd International Conference on Computer Science, Applied Mathematics and Applications (ICCSAMA 2014)}, pages 325--338. Springer.

\bibitem[{Lin(2004)}]{lin2004rouge}
Chin-Yew Lin. 2004.
\newblock Rouge: A package for automatic evaluation of summaries.
\newblock In \emph{Text summarization branches out}, pages 74--81.

\bibitem[{Mostow and Chen(2009)}]{mostow2009generating}
Jack Mostow and Wei Chen. 2009.
\newblock Generating instruction automatically for the reading strategy of self-questioning.
\newblock In \emph{AIED}, pages 465--472. Brighton.

\bibitem[{Papineni et~al.(2002)Papineni, Roukos, Ward, and Zhu}]{papineni2002bleu}
Kishore Papineni, Salim Roukos, Todd Ward, and Wei-Jing Zhu. 2002.
\newblock Bleu: a method for automatic evaluation of machine translation.
\newblock In \emph{Proceedings of the 40th annual meeting of the Association for Computational Linguistics}, pages 311--318.

\bibitem[{Rajpurkar et~al.(2016)Rajpurkar, Zhang, Lopyrev, and Liang}]{rajpurkar-etal-2016-squad}
Pranav Rajpurkar, Jian Zhang, Konstantin Lopyrev, and Percy Liang. 2016.
\newblock \href {https://doi.org/10.18653/v1/D16-1264} {{SQ}u{AD}: 100,000+ questions for machine comprehension of text}.
\newblock In \emph{Proceedings of the 2016 Conference on Empirical Methods in Natural Language Processing}, pages 2383--2392, Austin, Texas. Association for Computational Linguistics.

\bibitem[{Song et~al.(2018)Song, Wang, Hamza, Zhang, and Gildea}]{song-etal-2018-leveraging}
Linfeng Song, Zhiguo Wang, Wael Hamza, Yue Zhang, and Daniel Gildea. 2018.
\newblock \href {https://doi.org/10.18653/v1/N18-2090} {Leveraging context information for natural question generation}.
\newblock In \emph{Proceedings of the 2018 Conference of the North {A}merican Chapter of the Association for Computational Linguistics: Human Language Technologies, Volume 2 (Short Papers)}, pages 569--574, New Orleans, Louisiana. Association for Computational Linguistics.

\bibitem[{Taori et~al.(2023)Taori, Gulrajani, Zhang, Dubois, Li, Guestrin, Liang, and Hashimoto}]{alpaca}
Rohan Taori, Ishaan Gulrajani, Tianyi Zhang, Yann Dubois, Xuechen Li, Carlos Guestrin, Percy Liang, and Tatsunori~B. Hashimoto. 2023.
\newblock Stanford alpaca: An instruction-following llama model.
\newblock \url{https://github.com/tatsu-lab/stanford_alpaca}.

\bibitem[{Touvron et~al.(2023)Touvron, Lavril, Izacard, Martinet, Lachaux, Lacroix, Rozi{\`e}re, Goyal, Hambro, Azhar et~al.}]{touvron2023llama}
Hugo Touvron, Thibaut Lavril, Gautier Izacard, Xavier Martinet, Marie-Anne Lachaux, Timoth{\'e}e Lacroix, Baptiste Rozi{\`e}re, Naman Goyal, Eric Hambro, Faisal Azhar, et~al. 2023.
\newblock Llama: Open and efficient foundation language models.
\newblock \emph{arXiv preprint arXiv:2302.13971}.

\bibitem[{Zhang et~al.(2019)Zhang, Kishore, Wu, Weinberger, and Artzi}]{zhang2019bertscore}
Tianyi Zhang, Varsha Kishore, Felix Wu, Kilian~Q Weinberger, and Yoav Artzi. 2019.
\newblock Bertscore: Evaluating text generation with bert.
\newblock \emph{arXiv preprint arXiv:1904.09675}.

\end{thebibliography}

\appendix
\section{Details of Sem-Ent}
\label{sec:Details of Sem-Ent}
Inspired by the method proposed in the previous study\cite{han2022measuring}, we evaluate the semantic diversity of generated questions (Sem-Ent) from an entropy perspective. Let:

\begin{itemize}
\small
    \item $\mathcal{P}$ represent the set of different companies' presentations in our training data.
    \item $\mathcal{Q}_p$ denote the set of questions generated by the model for a given presentation $p \in \mathcal{P}$.
    \item $BERTopic_p$ be the BERTopic model trained specifically for company $p$.
    \item $\tilde{p}(j)$ represent the probability distribution of question $j$ across various topics, obtained from $BERTopic_p$.
    \item $k$ be the number of distinct semantic topics identified by $BERTopic_p$.
\end{itemize}

The Semantic Entropy (Sem-Ent) for a given Retriever-Model (RM) is calculated as:

\begin{equation}
    Sem\text{-}Ent(RM) = -\sum_{j=1}^{k} \tilde{p}(j) \cdot \log \tilde{p}(j)
\end{equation}

This metric, Sem-Ent, decreases when the semantic distribution becomes more imbalanced, indicating that models generate questions related to only a few specific semantic topics. Conversely, Sem-Ent reaches its maximum value of $\log k$ when generated questions are uniformly distributed across each semantic topic.

In our analysis, we observed that analysts often pose complex questions, comprising several sub-questions within a single paragraph. These include related follow-up questions or inquiries addressing different themes raised by various managers. To more finely assess the semantic diversity among these questions, we adopted a granular approach. For each generated question paragraph, we first performed sentence segmentation to calculate the topic probability distribution of each sentence. We then averaged these with the overall topic distribution of the entire question paragraph. Based on this aggregated distribution, we computed the Semantic Entropy (Sem-Ent) to gauge the diversity of the questions.

\section{Implement Details}
\label{sec:Implement Details}

Our system utilized the Alpaca-Lora-7B~\cite{alpaca, hu2021lora} as the question generator and the Guanaco-33B~\cite{dettmers2023qlora} for retrieving relevant passages. Both models, derivatives of the open-source LLaMA, were optimized using quantization and LoRA techniques. This allowed fine-tuning of only a small fraction of parameters: 0.24\% for the generator and 1.24\% for the retriever. We applied 8-bit and 4-bit quantization to the generator and retriever, respectively, and used gradient checkpointing for efficient VRAM utilization. The input length limits were set to 1400 tokens for the generator and 512 for the retriever, with the generator's output capped at 200 tokens.

During our experiments, the combined retriever-generator model was fine-tuned over three epochs using the AdamW optimizer, with a learning rate of \(2 \times 10^{-4}\) for both models. We employed a linear scheduler with a 0.1 warm-up ratio for learning rate adjustment, a maximum gradient norm of 1.0, and gradient accumulation over 32 steps, achieving an effective batch size of four. For inference, we used a top-p sampling strategy with \( p=0.9 \) and a temperature setting of 0.7. The complete training, inclusive of validation and model persistence, spanned roughly 38 hours on a singular NVIDIA A100 80GB GPU, consuming about 60GB of VRAM.
A pivotal hyperparameter in our setup was the top-k value, signifying the number of the most pertinent presentation passages that the retriever sends to the generator. Guided by the reference question, we consistently set top-k to 6 for all experiments.

\end{document}